\documentclass[a4paper,twoside]{article}

\usepackage{epsfig}
\usepackage{subcaption}
\usepackage{calc}
\usepackage{dsfont}
\usepackage{amsmath}
\usepackage{amsthm}
\usepackage{siunitx}
\usepackage{multicol}
\usepackage{pslatex}
\usepackage{algorithm2e}
\usepackage[bottom]{footmisc}
\usepackage{url}
\usepackage{acro}
\usepackage [utopia]{mathdesign}
\usepackage [OMLmathrm,OMLmathbf,sfdefault=fav,scaled=0.875]{isomath}
\usepackage[defaultcolor=magenta]{changes}
\usepackage{booktabs}
\usepackage{makecell}
\usepackage{multicol}
\usepackage{ragged2e}
\usepackage{hyperref}
\usepackage{nicefrac}
\usepackage{SCITEPRESS}     % Please add other packages that you may need BEFORE the SCITEPRESS.sty package.

\usepackage{sectsty}% http://ctan.org/pkg/sectsty.
\sectionfont{\MakeUppercase}

\robustify\bfseries
\usepackage{apalike}
\renewcommand{\vec}{\vectorsym}
\newcommand{\mat}{\matrixsym}

\DeclareAcronym{bve}{short = BVE, long = best viewpoint estimator}
\DeclareAcronym{ekf}{short = EKF, long = extended Kalman filter}
\DeclareAcronym{3d}{short = 3D, long = three-dimensional}
\DeclareAcronym{6d}{short = 6D, long = six-dimensional}
\DeclareAcronym{mape}{short = MAPE, long = mean absolute percentage error}
\DeclareAcronym{mae}{short = MAE, long = mean absolute error}
\DeclareAcronym{rmse}{short = RMSE, long = root mean square error}
\DeclareAcronym{mse}{short = MSE, long = mean square error}
\DeclareAcronym{dof}{short = DoF, short-plural = , long = Degree of freedom, plural-form = Degrees of freedom, first-style=footnote}
\DeclareAcronym{un}{short = UN, long = United Nations}
\DeclareAcronym{ods}{short = ODS, long = objectives for sustainable development}
\DeclareAcronym{rgb-d}{short = RGB-D, long = {Red, green, blue and depth sensor} , first-style = footnote}

\begin{document}

\title{BVE + EKF: A viewpoint estimator for the estimation of the object's position in the 3D task space using Extended Kalman Filters}

\ifdefined\DOUBLEBLIND\else
\author{\authorname{\normalsize Sandro Costa Magalhães\sup{1,2}\orcidAuthor{0000-0002-3095-197X}, António Paulo Moreira\sup{1,2}\orcidAuthor{0000-0001-8573-3147}, Filipe Neves dos Santos\sup{2}\orcidAuthor{0000-0002-8486-6113} and Jorge Dias\sup{3,4}\orcidAuthor{0000-0002-2725-8867}}
\begin{multicols}{2}
\affiliation{\sup{1}INESC TEC, Porto, Portugal}
\affiliation{\sup{2}FEUP, Porto, Portugal}
\affiliation{\sup{3}ISR, University of Coimbra, Coimbra, Portugal}
\affiliation{\sup{4}KUCARS, Khalifa University, Abu Dhabi, UAE}
\end{multicols}
\email{\{sandro.a.magalhaes, filipe.n.santos\}@inesctec.pt, amoreira@fe.up.pt, jorge.dias@ku.ac.ae}
}
\fi


% \quickwordcount{Example}
% \quickcharcount{Example}
% \detailtexcount{Example}

\abstract{
\Acs{rgb-d} sensors face multiple challenges operating under open-field environments because of their sensitivity to external perturbations such as radiation or rain. Multiple works are approaching the challenge of perceiving the \ac{3d} position of objects using monocular cameras. However, most of these works focus mainly on deep learning-based solutions, which are complex, data-driven, and difficult to predict. So, we aim to approach the problem of predicting the \ac{3d} objects' position using a Gaussian viewpoint estimator named \ac{bve}, powered by an \ac{ekf}. The algorithm proved efficient on the tasks and reached a maximum average Euclidean error of about \qty{32}{\milli\meter}. The experiments were deployed and evaluated in MATLAB using artificial Gaussian noise. Future work aims to implement the system in a robotic system. 
}

\onecolumn \maketitle \normalsize \setcounter{footnote}{0} \vfill

\section{Introduction}
\label{sec:introduction}

Agriculture is a critical sector that has been facing several difficulties over time. That constraints are well-designed by several organizations, such as the \ac{un} in the \ac{ods} \cite{GeneralAssembly2015}. However, their solution is still a challenge. 

The increased food demand promoted by the population growth \cite{FAOUS2023} and labor shortage require improved and efficient agricultural technologies that may speed up the execution of farming tasks. Monitoring and harvesting are some tasks that may benefit from robotization in the area. 

Autonomous robots require robust perception systems to detect and identify fruits and other agricultural objects. Several works use \ac{rgb-d} cameras to see the objects and infer their \ac{3d} position \cite{Kumar2022,Magalhaes2022a}. However, \ac{rgb-d} sensor can malfunction under open-field environments due to external interferences \cite{Kumar2022,GeneMola2020,Ringdahl2019}, such as radiation or rain. To overcome this challenge, several works designed solutions to infer the position of the objects from monocular sensors. The most common systems are based on deep learning to infer the \ac{6d} or \ac{3d} pose of objects \cite{Li2023,Parisotto2023,Chang2021,Wang2021} or estimate their depth \cite{Ma2019,Birkl2023}. Deep learning deploys, although complex, algorithms that are very data-dependent, usually supervised, and hard to predict and modify.

Despite the tendency for deep-learning solutions, we still can use Bayesian algorithms to infer the \ac{3d} position of objects. In previous work, \cite{Magalhaes2024} designed the MonoVisual3DFilter to estimate the position of objects using histogram filters. However, the algorithm still requires the manual definition of viewpoints to estimate the position of the fruits.

Therefore, in this work, we challenge to identify a mechanism that can autonomously infer the \ac{3d} position of fruits without the demand for manually defining viewpoints. 

We approach our question with the challenge of autonomously identifying the position of fruits, such as tomatoes, in the tomato plant for precision monitoring or harvesting tasks. We assume the system has a manipulator with a monocular camera configured in an eye-hand manner.  

In the following sections, this article explores the proposed solution. The section \ref{sec:materials and methods} details the implementation of the \ac{bve} powered by the \ac{ekf} to estimate the \ac{3d} position of the objects in the tasks space. This section also defines some experiments to evaluate the algorithm. The section \ref{sec:results} illustrates the results for the various experiments and some algorithm limitations. Finally, section \ref{sec:conclusion} concludes and summarizes this study and introduces future work. 

\section{Materials and Methods}
\label{sec:materials and methods}

\subsection{\Ac{bve}}
\label{subsec: bve}

A statistical optimization approach guides towards a solution for this problem. The observation of a fruit from a viewpoint has an associated observation error. The goal is to identify a subsequent viewpoint that minimizes this error. Thus, the problem is the minimization of a loss function related to the intersection of Gaussians distributions \eqref{eq: classical product of gaussians}, where \(N_i(\vec{\mu}_i, \mathbf{\Sigma}_i)\) denotes a Gaussian distribution. The index \(i \in \mathds{N}\) corresponds to each observation viewpoint.

\begin{equation}
    {N}({\vec{\mu_p}}, \mat{\Sigma_p}) = {N}_1({\vec{\mu}}, \mat{\Sigma_1}) \cdot \ldots \cdot {N}_n({\vec{\mu}}, \mat{\Sigma_n}) \label{eq: classical product of gaussians}
\end{equation}

The Gaussian distribution's product \eqref{eq: classical product of gaussians} presents significant computational complexity. Nevertheless, \cite{Petersen2012} posits that we can decompose the product of Gaussians into two distinct equations---addressing the mean values and the covariance. Because we expect the fruit to remain stopped while hung on the tree, this solution proposes that the position of the tomato, \(\vec{k}\), remains invariant, thus \(\vec{\mu}_i = \vec{k}\). Hence, we simplify the optimization problem to \eqref{eq: soma covariancias}.

\begin{equation}
    \mat{\Sigma_p} = (\mat{\Sigma_1}^{-1} + \ldots + \mat{\Sigma_n}^{-1})^{-1}
    \label{eq: soma covariancias}
\end{equation}

The observation noise covariance is predominantly a characteristic intrinsic to the camera. Consequently, the camera's covariance \(\mathbf{\Sigma}_c\) remains constant within the camera's frame, \(C\). The movement of the camera to different poses, $\Vec{c}$, changes the observation noise in the fixed world frame \(W\). So, the model requires an observation covariance matrix expressed within the main frame \(W\) \eqref{eq: rotacao covariancia} to correlate the multiple observations. The matrix \(\mathbf{R_C^W}\) represents a rotation matrix that delineates the relationship between the camera's frame \(C\) and the main frame \(W\).

\begin{equation}
    \mat{\Sigma_n} = \mat{R_C^W} \mat{\Sigma_c} \mat{R_C^W}^\intercal
    \label{eq: rotacao covariancia}
\end{equation}

In concluding the initial problem definition, we should recognize that the covariance matrix undergoes modification with each iteration of the algorithm as a consequence of the computations performed in equations \eqref{eq: soma covariancias} and \eqref{eq: rotacao covariancia}. To generalize the system's initial conditions, a generic covariance matrix, \(\mathbf{\Sigma}_o\), is considered. This matrix represents the culmination of all previous intersections of covariance matrices up to the point \(k-1\).

\subsubsection{Definition of the rotation matrix}
\label{subsubsec: rotation matrix}

\begin{figure}[!htb]
    \centering
    \includegraphics[width=0.7\linewidth]{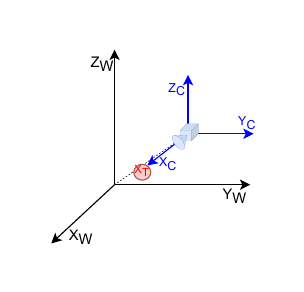}
    \caption{Definition of the camera's frame and the main frame.}
    \label{fig: frames}
\end{figure}

The observation covariance matrix $\mat{\Sigma_c}$ is initially defined into the camera's frame. We can convert data between frames using homogeneous transformations, namely rotation matrices, because the translation is irrelevant. Figure \ref{fig: frames} illustrates a possible generic relationship between frames. The camera's frame, denoted as $O{x_C y_C z_C}$, is centered at the sensor, and the $\vec{x_C}$ axis indicates the camera's viewing direction. For simplicity, we assume that $\vec{y_C}$ is always parallel to the plane defined by $_{x_W} O_{y_W}$. This simplification is possible because the covariance matrix is ideally symmetrical in the $\vec{x_C}$ axis, and the other axis's orientation is irrelevant.

Given the fruit position's estimation in the main frame, $\vec{\hat{k}}$, $\vec{e_{x_C^W}}$ denotes the unit vector of $\vec{x_C}$ axis \eqref{eq: e_x_c^W}, where $\vec{c}$ is the camera's position. We can define the camera frame's axes in the main frame through \eqref{eq: X_C^W}, \eqref{eq: Y_C^W}, and \eqref{eq: Z_C^W}. The rotation matrix $\mat{R_C^W}$ that relates the camera frame to the main frame is obtained from \eqref{eq: R_C^W}. In \eqref{eq: R_C^W}, $\vec{e_{x_C^W}}$, $\vec{e_{y_C^W}}$, and $\vec{e_{z_C^W}}$ are the unit vector of $\vec{{x_C^W}}$, $\vec{{y_C^W}}$, and $\vec{{z_C^W}}$, respectively.

{\allowdisplaybreaks
\begin{align}
    \vec{e_{x_c^W}} = & \dfrac{\vec{\hat{k}} - \vec{c}}{||\vec{\hat{k}} - \vec{c}||} \label{eq: e_x_c^W}\\
    \vec{x_C^W} = & \vec{e_{x_c^W}} = \begin{bmatrix} x_1 & x_2 & x_3 \end{bmatrix}^\intercal \label{eq: X_C^W} \\
    \vec{y_C^W} = & \begin{bmatrix} -x_2 & x_1 & 0 \end{bmatrix}^\intercal \label{eq: Y_C^W}\\
    \vec{z_C^W} = & \vec{x_C^W} \times \vec{y_C^W} %= \begin{bmatrix}
    %    -x_1 \times x_3 \\ -x_2 \times x_3 \\ x_1^2 + x_2^2
    %\end{bmatrix} 
    \label{eq: Z_C^W}\\
    \mat{R_C^W} = & \begin{bmatrix}\vec{e_{x_C^W}} & \vec{e_{y_C^W}} & \vec{e_{z_C^W}} \end{bmatrix} \label{eq: R_C^W}
\end{align}
}

\subsubsection{Definition of the objective function}
\label{subsubsec: objective function}

We aim to minimize a function related to the product of Gaussian distributions \eqref{eq: soma covariancias}. This endeavor requires a loss function directly contingent upon the Gaussian intersection. The optimizer predicates a scalar output from the loss function we designed as a dispersion dependency.

For each observation, \eqref{eq: sigma updated} characterizes the intersection of two covariance matrices in a fixed main frame. The computation of this intersection necessitates the calculation of three inverse matrices, which is a computationally demanding operation.

\begin{equation}
    \mat{\Sigma_{u}} = (\mat{\Sigma_o}^{-1} + \mat{\Sigma_n}^{-1})^{-1}
    \label{eq: sigma updated}
\end{equation}

We reduced the number of these operations through the precision matrix ($\mat{P} = \mat{\Sigma_u}^{-1}$). Then, the precision's concentration ($c = \det(\mat(P))$) translates the matrix into a scalar. So, we can define the objective function as the dispersion ($\nicefrac{1}{c}$), because, according to the properties of the determinants, $\det(\mat{P}^{-1}) = \det(\mat{P})^{-1}$. Due to the low magnitude of the loss function, we scaled the dispersion into the logarithmic scale \eqref{eq: loss function}. $\mat{P}$ and $\mat{\Sigma_n}$ are dependent on $\vec{\hat{c}}$, the next estimated position of the camera, which we aim to optimize. 

\begin{align}
    f(\vec{\hat{c}}) = &  -\ln(\det(\mat{\Sigma_n}^{-1} + \mat{\Sigma_o}^{-1})) \label{eq: loss function}
\end{align}

Alternatively to the loss function \ref{eq: loss function}, we can minimize the absolute maximum eigen value of the covariance matrix if we have enough computing power to compute \eqref{eq: sigma updated}. While using this loss function, we should remember that it is highly non-linear and whose derivative function varies at each step because of the maximum function.

\begin{equation}
    f(\vec{\hat{c}}) = \max( | \text{eig}(\mat{\Sigma_u}) | )
    \label{eq: loss function v2}
\end{equation}

We can use optimization algorithms operating with non-linear restrictions and loss functions to solve the problem using both functions. For the current analysis, we opted to use an interior-point algorithm \cite{Nocedal2014}, already implemented in MATLAB's optimization toolbox\cite{MathWorks2024}.

%We will observe later that \eqref{eq: loss function} can effectively estimate the next best viewpoint is empirically faster to compute. However, the \eqref{eq: loss function v2} tends to deliver slightly better results. 

We also intend to effectively drive the camera to the objects to perform tasks, while estimating the position of the fruit. Towards that, we added an additional component to the loss function \eqref{eq: improved loss function}. The $\text{act}(i, a, b)$ is a sigmoid activation function \eqref{eq: sigmoid}. The sigmoid activates the additional component, forcing the camera to approximate the object. In the activation function \eqref{eq: sigmoid}, $a$ and $b$ are control parameters that set its aggressiveness and its set point (i.e., the value of the function for $act(\cdot) = 0.5$), respectively; $i$ is the iteration number of the procedure. Through this strategy, we can activate gradually the Euclidean error to the fruit according to the evolution of the estimation procedure. 

\begin{align}
    \text{act}(i, a, b) &= \dfrac{1}{1 + \text{e}^{-a \cdot (i - c)}} \label{eq: sigmoid} \\
    F(\vec{\hat{c}}) &= f(\vec{\hat{c}}) + \text{act}(i, a, b) \cdot ||\vec{\hat{k}} - \vec{\hat{c}}|| \label{eq: improved loss function}
\end{align}

\subsubsection{Definition of the restrictions}
\label{subsubsec: restrictions}

%We should constrain the optimizer to match the environment and hardware constraints to estimate the fruit's position under real conditions objectively.
The proposed algorithm can effectively estimate the best camera poses that maximize the observability of the fruits. However, some restrictions should be implemented to match the environment and hardware constraints. 
So, we defined that the selected poses must be inside the working space of a 6 \acp{dof} manipulator. A spheric model simplifies this workability restriction. Considering a manipulator with a working space centered in $\vec{m}$ and with a radius $r_m$, in meters, the camera's pose $\vec{\hat{c}}$ must be inside \eqref{eq: restricao area manipulador}. We only estimate the center position of the fruit but mislead its volume. An additional condition forces the camera to be outside the fruit space. Thus, considering an average fruit radius $r_k$, centered in $\vec{k}$, the camera's pose has to comply with \eqref{eq: restricao tomate}.

\begin{align}
    (\vec{\hat{c}} - \vec{m}) \cdot (\vec{\hat{c}} - \vec{m})^\intercal - r_m^2 & \leq 0 \label{eq: restricao area manipulador}  \\
    - (\vec{\hat{c}} - \vec{\hat{k}}) \cdot (\vec{\hat{c}} - \vec{\hat{k}})^\intercal + r_k^2 & \leq 0
    \label{eq: restricao tomate}
\end{align}

\begin{figure}[!htb]
    \centering
    \includegraphics[width=0.5\linewidth]{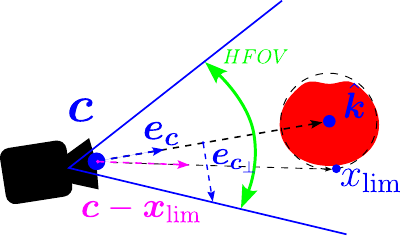}
    \caption{Restriction---fruit inside the camera's field of view}
    \label{fig:cone-view}
\end{figure}

The camera's orientation is also relevant to ensure it looks towards the fruit. The algorithm only focuses in estimating the best position for the camera, but also the orientation of it should be constrained, ensuring the camera is looking towards the fruits. The camera has a conical vision profile. So, we constrained the fruits to be inside the camera's field of view, with a conical shape, \eqref{eq: restricao visao conica da camara} and figure \ref{fig:cone-view}, where $HFOV$ is the angle of the horizontal field of view of the camera in radians. 

{\allowdisplaybreaks
\begin{align}
    \vec{e_{c}} = & \frac{\vec{\hat{k}} - \vec{\hat{c}}}{||\vec{\hat{k}} - \vec{\hat{c}}||} \\
    \vec{e_{c_\perp}} = & \begin{bmatrix} - e_{c,2} & e_{c,1} & e_{c,3} \end{bmatrix}^\intercal \\
    \vec{x_{\mathbf{lim}}} = & \vec{\hat{k}} + r_k \cdot \vec{e_{c_\perp}} \\
    0 \geq & \dfrac{\vec{x_{\mathbf{lim}}} - \vec{\hat{c}}}{||\vec{x_{\mathbf{lim}}} - \vec{\hat{c}}||} \cdot \vec{e_{c}} - \cos\left(\dfrac{HFOV}{2}\right) \label{eq: restricao visao conica da camara}
\end{align}
}

\begin{figure}[!htb]
    \centering
    \includegraphics[width=0.7\linewidth]{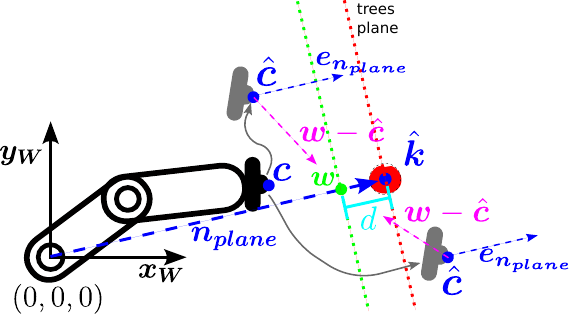}
    \caption{Restriction---camera cannot cross the fruits' trees}
    \label{fig:plane}
\end{figure}

In a tomato greenhouse, where plants are aligned in rows, the robot must avoid crossing these rows to prevent damage. These rows make a wall of uncrossable tomato plants. Avoiding crossing the designed walls is managed by defining a restriction \eqref{eq: restricao parede de plantas}, modelling the plant rows as a planar boundary to keep the robot on one side, set at a distance $d$ meters from the fruit, as illustrated in the figure \ref{fig:plane}. The plane's orientation is determined by the normal unit vector $\vec{e_{n_\textbf{plane}}}$, which represents the normal vector $\vec{n_\textbf{plane}}$.

\begin{align}
    \vec{n_\mathbf{plane}} = & \begin{bmatrix}\hat{k}_0 & \hat{k}_1 & 0 \end{bmatrix}^\intercal \label{eq: normal vector}\\
    \vec{w} = & \vec{\hat{k}} - d \cdot \vec{e_{n_\mathbf{plane}}} \label{eq: wall} \\
    0 \geq & \vec{e_{n_\mathbf{plane}}} \cdot (\vec{\hat{c}} - \vec{w})\label{eq: restricao parede de plantas}
\end{align}

In addition to the previous restrictions, we designed extra ones based on the manipulator's specific features. These ensure that only valid poses are selected, making the kinematics computable, which varies with the manipulator's kinematics. 

The previously designed constraints are very specific and complex for the designed target. Conducting essays with simplified algorithms should be advantageous in understanding the benefits and the limitations of a more complete scene design.  So, we also conducted experiments with simplified restrictions, considering just one: the distance between the camera and the fruit, denoted as $l_{dist}$ in \eqref{eq: restriction dist}.

\begin{equation}
    l_{dist} - \varepsilon < ||\vec{\hat{c}} - \vec{\hat{k}}|| <  l_{dist} + \varepsilon \label{eq: restriction dist}
\end{equation}

\subsection{Fruit pose estimation using the \ac{ekf}}
\label{subsec: ekf}

The \ac{bve} computes the best observability viewpoints but does not estimate the \ac{3d} position of the objects. Based on an initial rough estimation of the position of the fruit, the \ac{ekf} can provide iterative refinement of the objects' position.

To ensure a good correct operation of the \ac{ekf}, the camera should move smoothly while looking at the fruit to correct the fruit position estimation iteratively. So, an additional restriction is implemented to the \ac{bve} to ensure that the camera moves to the next best viewpoint in a radius of $r_d$ meters, \eqref{eq: max mov}. 

\begin{equation}
    ||\vec{\hat{c}_{k+1}} - \vec{c_{k}}|| - r_d < 0
    \label{eq: max mov}
\end{equation}

% The \ac{ekf} is divided into two main steps (figure \ref{fig: ekf}): the prediction phase and the correction phase. The fruit position is continuously estimated during prediction, attending dynamics and predictive movement. At the correction phase, the fruit is observed by a dedicated sensor, and so its position is corrected according to the measurements performed. 
The \ac{ekf} is divided into two main steps: the prediction phase and the correction phase. The fruit position is continuously estimated during prediction, attending dynamics and predictive movement. At the correction phase, the fruit is observed by a dedicated sensor, and so its position is corrected according to the measurements performed.

% \begin{figure}[!htb]
%     \centering
%     \includegraphics[width=\linewidth]{figures/EKF_diagram.pdf}
%     \caption{Diagram of the \ac{ekf} applied.}
%     \label{fig: ekf}
% \end{figure}

\paragraph{Prediction} 
%\label{subsubsec: ekf-prediction}

During the prediction phase, we estimate the fruit's position, attending to its zero dynamics. The \ac{ekf} should expect the fruit to not move. So, the predicted position of the fruit is the same as the previous one \eqref{eq: estimated_x}. Besides, the \ac{ekf} also has to propagate the covariance estimation error \eqref{eq: P propagation}.

{\allowdisplaybreaks
\begin{align}
    \vec{\hat{x}_{k|k-1}} = & f(\vec{\hat{x}_{k-1|k-1}}, \vec{u_{k-1}}) = \mat{I} \cdot \vec{\hat{x}_{k-1}} \label{eq: estimated_x} \\ 
    \mat{P_{k|k-1}} = & \mat{F_k} \cdot \mat{P_{k|k-1}} \cdot \mat{F_k^\intercal} + \mat{Q_{k}} = \mat{P_{k|k-1}} + \mat{Q_{k}} \label{eq: P propagation}\\
    \mat{F_k} = & \dfrac{\partial f}{\partial \vec{x}}\biggr\rvert_{\vec{\hat{x}_{k-1|k-1}, \vec{u_k}}} = 1 
\end{align}
}

\paragraph{Correction}
%\label{subsubsec: ekf-correction}

Assuming that the camera always observes the fruit, the \ac{ekf} always has a correction phase after a prediction phase. During this phase, the \ac{ekf} corrects the estimation of the fruit's position \eqref{eq: x correction}, acknowledging the measures obtained from the camera to the sensor \eqref{eq: measure}. The \ac{ekf} uses the same rough initial estimation method based on the fruit's average size and the camera's distortion. The correction phase also corrects the covariance propagation error \eqref{eq: P correction}. In these equations, $\vec{\hat{x}}$ is the estimated position of the fruit for each instance, and $\varepsilon$ is a random noise variable added to create noise for the simulated environment (under real conditions, this value is realistically measured and should be ignored).

{\allowdisplaybreaks
\begin{align}
    h(\vec{\hat{x}_{k|k-1}}) = & ||\vec{\hat{x}_{k-1}} - \vec{c}|| \label{eq: measure} \\
    \vec{z_k} = & ||\vec{k} - \vec{c} || + \varepsilon \cdot \sqrt{\sigma_{xx}} \\
    \mat{H_k} = & \nabla h(\vec{\hat{x}_{k|k-1}}) = \dfrac{\vec{\hat{x}_{k-1}}-\vec{c}}{||\vec{\hat{x}_{k-1}} - \vec{c}||} \\
    \mat{K_k} = & \mat{P_{k|k-1}} \cdot \mat{H_k}^\intercal \cdot (\mat{H_k} \cdot \mat{P_{k|k-1}} \cdot \mat{H_{k}}^\intercal + {R_k})^{-1} \\
    {R_k} = & \sigma_{xx} \\
    \mat{P_{k|k}} = & (\mat{I} - \mat{K_k} \cdot \mat{H_k}) \cdot \mat{P_{k|k-1}} \label{eq: P correction}\\
    \vec{\tilde{y}_k} = & \vec{z_k} - h(\vec{\hat{x}_{k|k-1}}) \\
    \vec{\hat{x}_{k|k}} = & \vec{\hat{x}_{k|k-1}} + \mat{K_k} \cdot \vec{\tilde{y}_k} \label{eq: x correction}
\end{align}
}

\subsection{Experiments}
\label{sec: bve/experiments}

Multiple essays were made under a simulation context in MATLAB to validate the designed algorithms. We deployed an iterative protocol that adds restrictions to the optimizer. That helps to understand the limitations with the increase of the optimization difficulties. Bellow, we systematize the different experiments and the restrictions considered for each one. Mind that in all cases, the \ac{bve} always considers the restriction \eqref{eq: max mov} in $r_d$ of \qty{0.2}{\meter}. The \ac{ekf} uses a near-realistic covariance matrix for the camera's observations of the fruit, corresponding to a diagonal matrix and a bigger observation variance in the $xx$ axis. 

% For the different restrictions and experiments, we consider the followin that the camera should be in a radius $l_{dist}$ of \qty{1 +- 0.1}{\meter} to the fruit. Concerning the other constraints, $d = 0.1$~\si{\metre}; the considered manipulator is the Robotis Manipulator-H, with a working space centred in $\vec{m} =  \begin{bmatrix}0 & 0 & 0.159\end{bmatrix}^\intercal$~\si{\metre} and radius $r_m = 0.645$~\si{\metre}. A near-realistic covariance matrix for the observations of the camera to the fruit, that corresponds to a diagonal matrix and a bigger observation variance in the $xx$ axis. 

% Further, we added the \ac{ekf} algorithm to the \ac{bve}, to continuously observe and correct the position of the fruit in the \ac{3d} task space. Similar essays were made to evaluate the performance of the algorithm. For this case, we also essay both designed loss functions. In the \ac{ekf}, we consider a maximum step size $r_d = 0.2$~\si{\metre}, applied to all the experiments. 

% To systematize the different experiments, we define the protocol below.

\begin{description}
    \item[E1] For this experiment, we used the loss function \eqref{eq: loss function} and restricted the \ac{bve}'s behavior with limited the position of the camera $l_{dist}$ of \qty{1+-0.1}{\meter} to the fruit, \eqref{eq: restriction dist}.
    \item[E2] In this experiment, we repeated the previous essay, but we also considered the restriction \eqref{eq: restricao area manipulador}  that assures that the camera is inside the manipulator's working space, considering the Robotis Manipulator-H with its center $\vec{m}$ in $\begin{bmatrix}0 & 0 & 0.159\end{bmatrix}^\intercal$~\unit{\meter} and a maximum range $r_m$ of \qty{0.645}{\meter}.
    \item[E3] In this experiment, we consider the loss function \eqref{eq: loss function} and the restrictions \eqref{eq: restricao area manipulador}, \eqref{eq: restricao tomate} with the average tomato size $r_k$, and \eqref{eq: restricao visao conica da camara}. 
    \item[E4] This experiment considers the restrictions and the loss function of E3 and adds the restriction \eqref{eq: restricao parede de plantas}, considering $d = \qty{0.1}{\m}$;
    \item[E5] This experiment repeats the previous experiment, adding the kinematics constraints, ensuring that the camera's pose is always a valid pose for the manipulator;
    \item[E6] Repeats the experiment E1, considering the loss function \eqref{eq: loss function v2}, based on the minimization of the maximum covariance, instead of the dispersion-based loss function \eqref{eq: loss function};
    \item[E7] Repeats the experiment E2, considering the loss function \eqref{eq: loss function v2};
    \item[E8] Repeats the experiment E3, considering the loss function \eqref{eq: loss function v2};
    \item[E9] Repeats the experiment E4, considering the loss function \eqref{eq: loss function v2}; and
    \item[E10] Repeats the experiment E5, considering the loss function \eqref{eq: loss function v2}.
\end{description}

We executed the simulation code for 100 runs for each of these experiments. In each run, we consider a random position for the tomato $\vec{k}, k_i \in [-1, 1]$\,\unit{\meter}, and a random initial position for the camera $\vec{c}, c_i \in [-2, 2]$\,\unit{\meter}. The initial estimation of the fruit was initialized in its real position $\vec{k}$ added by a random bias between $[-0.15; 0.15]$\,\unit{\meter} for each axis. 

We assessed the algorithm's performance using the \ac{mape} \eqref{eq: mape}, \ac{mae} \eqref{eq: mae}, \ac{rmse} \eqref{eq: rmse}, and \ac{mse} \eqref{eq: mse}.

{\allowdisplaybreaks
\begin{align}
    \text{MAPE }(\mu_j, \hat{\mu}_j) = & \dfrac{1}{N\cdot M} \sum_i^N \sum_j^M \left|\dfrac{\mu_{ij}-\hat{\mu}_{ij}}{\mu_{ij}}\right| \times 100 \nonumber \\ & \qquad \forall j \in \mathds{N}:\{1 .. M\} \label{eq: mape} \\
    \text{MAE }(\mu_j,\hat{\mu}_j) = & \dfrac{1}{N\cdot M} \sum_i^N \sum_j^M |\mu_{ij} - \hat{\mu}_{ij}| \nonumber \\ & \qquad \forall j \in \mathds{N}:\{1 .. M\}\label{eq: mae} \\
    \text{MSE }(\mu_j,\hat{\mu}_j) = & \dfrac{1}{N\cdot M} \sum_i^N \sum_j^M (\mu_{ij} - \hat{\mu}_{ij})^2 \nonumber \\ & \qquad \forall j \in \mathds{N}:\{1 .. M\} \label{eq: mse} \\
    \text{RMSE }(\mu_j,\hat{\mu}_j) = & \sqrt{ \dfrac{1}{N\cdot M} \sum_i^N \sum_j^M (\mu_{ij} - \hat{\mu}_{ij})^2} \nonumber \\ & \qquad \forall j \in \mathds{N}:\{1 .. M\} \label{eq: rmse} 
\end{align}
}

\section{Results and discussion}
\label{sec:results}

\begin{figure*}[!t]
    \centering\hfill
    % \begin{subfigure}[b]{\linewidth}
    %     \centering
    %     \includegraphics[width=\linewidth]{figures/E2_1.eps}
    %     \caption{E2.1}
    %     \label{fig:E2.1}
    % \end{subfigure}\hfill
    % \begin{subfigure}[b]{0.49\textwidth}
    %     \centering
    %     \includegraphics[width=\textwidth]{figures/E2_6.eps}
    %     \caption{E2.6}
    %     \label{fig:E2.6}
    % \end{subfigure}\hfill
    \begin{subfigure}[b]{0.49\linewidth}
        \centering
        \includegraphics[width=\textwidth]{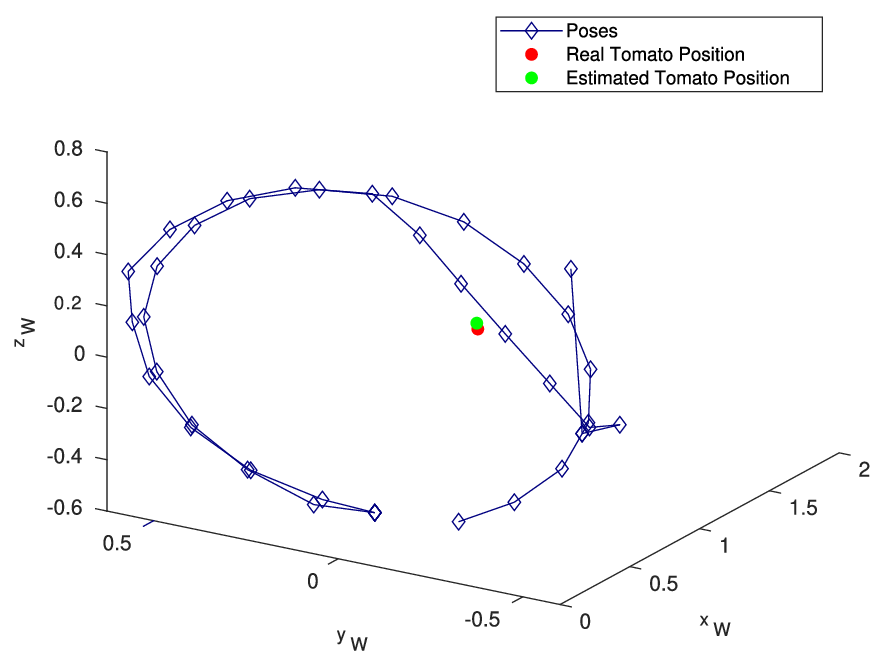}
        \caption{E2}
        \label{fig:E2.2}
    \end{subfigure}\hfill
    \begin{subfigure}[b]{0.49\linewidth}
        \centering
        \includegraphics[width=\textwidth]{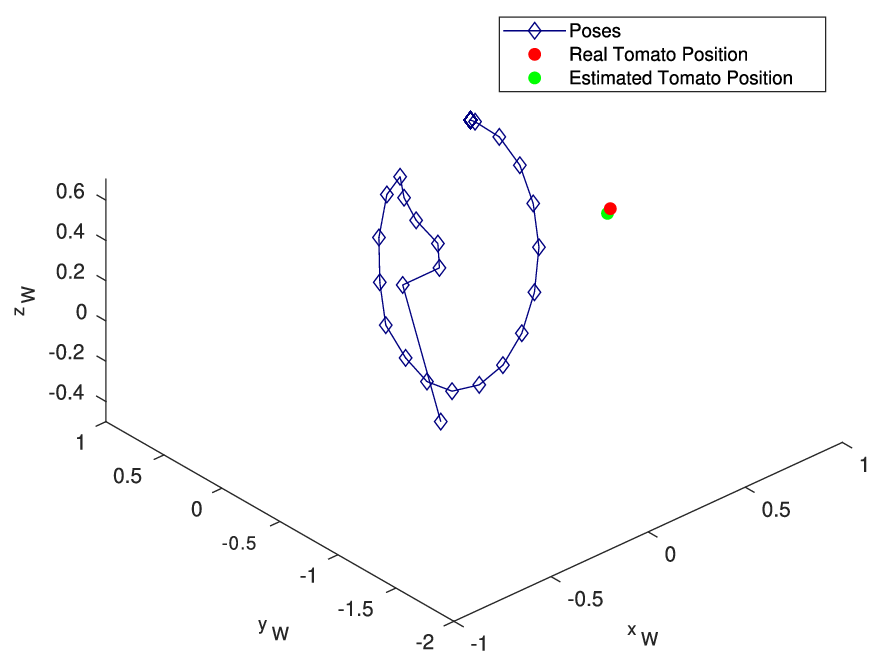}
        \caption{E7}
        \label{fig:E2.7}
    \end{subfigure}\hfill
    % \begin{subfigure}[b]{0.49\textwidth}
%         \centering
%         \includegraphics[width=\textwidth]{figures/E2_3.eps}
%         \caption{E2.3}
%         \label{fig:E2.3}
%     \end{subfigure}\hfill
%     \begin{subfigure}[b]{0.49\textwidth}
%         \centering
%         \includegraphics[width=\textwidth]{figures/E2_8.eps}
%         \caption{E2.8}
%         \label{fig:E2.8}
%     \end{subfigure}\hfill
% \end{figure}
%
% \begin{figure}[!htb]\ContinuedFloat
%     \centering
%     \begin{subfigure}[b]{0.49\textwidth}
%         \centering
%         \includegraphics[width=\textwidth]{figures/E2_4.eps}
%         \caption{E2.4}
%         \label{fig:E2.4}
%     \end{subfigure}\hfill
%     \begin{subfigure}[b]{0.49\textwidth}
%         \centering
%         \includegraphics[width=\textwidth]{figures/E2_9.eps}
%         \caption{E2.9}
%         \label{fig:E2.9}
%     \end{subfigure}\hfill
    \begin{subfigure}[b]{0.49\linewidth}
        \centering
        \includegraphics[width=\linewidth]{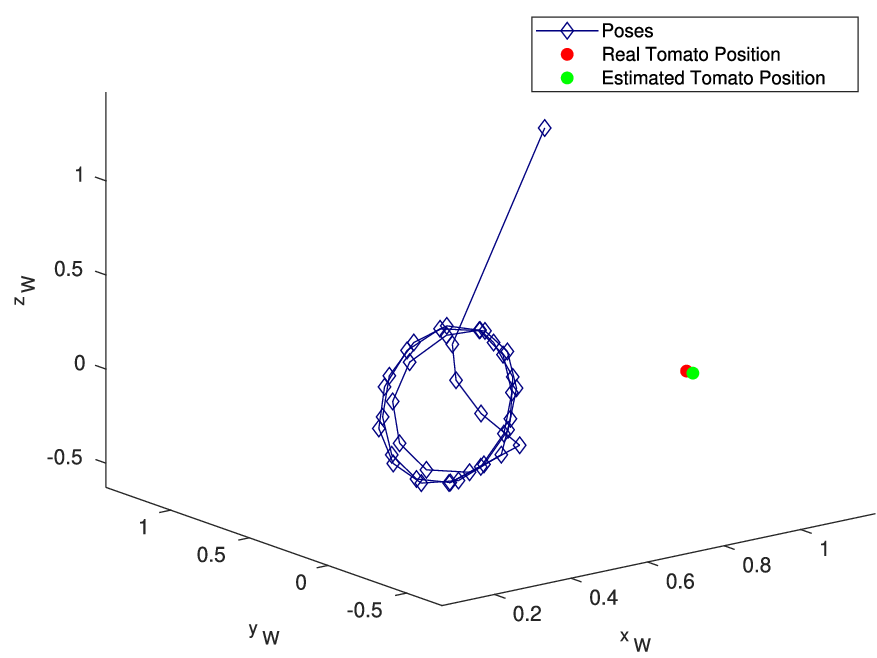}
        \caption{E5}
        \label{fig:E2.5}
    \end{subfigure}\hfill
    \begin{subfigure}[b]{0.49\linewidth}
        \centering
        \includegraphics[width=\linewidth]{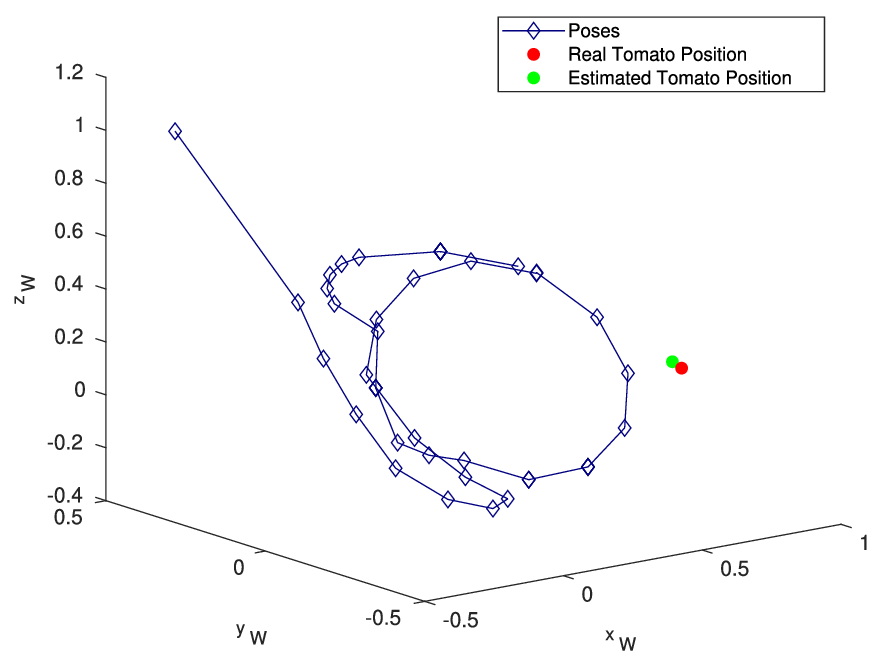}
        \caption{E10}
        \label{fig:E2.10}
    \end{subfigure}\hfill
    \caption{Sample paths generated by the different experiments to assess the fruit's position. Light blue -- poses; red -- real fruit position; green -- estimated fruit position.}
    \label{fig: bve+ekf motion plot}
\end{figure*}

%The \ac{bve} powered by the \ac{ekf} is a competitive or complementary strategy to MonoVisual3DFilter. 
The \ac{bve} powered with \ac{ekf} can effectively estimate the fruits' position while using a monocular camera. We organized ten experiments, as described in the section \ref{sec: bve/experiments}. Table \ref{tab: error bve + ekf} reports the average errors for the different experiments. Figure \ref{fig: bve+ekf motion plot} illustrates sample paths produced by the optimizer for experiments E2, E5, E7, and E10.

\begin{table}[!htb]
\centering
\caption{Error computations to the centre estimation using the \ac{bve} and the \ac{ekf}}
\label{tab: error bve + ekf}
\begin{tabular}{@{}lS[table-auto-round]S[table-auto-round]S[table-auto-round]S[table-auto-round]@{}}
\toprule
\textbf{} &
  {\thead{\ac{mape} \\ (\unit{\percent})}} &
  {\thead{\ac{mae}  \\ (\unit{\milli\metre})}} &
  {\thead{\ac{rmse} \\ (\unit{\milli\metre})}} &
  {\thead{\ac{mse}  \\ \unit{(\milli\metre)\squared}}} \\ \midrule
\textbf{E1}  & 5.12  & 37.1 & 15.5 & 241.64 \\
\textbf{E2}  & 12.86 & 52.5 & 25.5 & 650.47 \\
\textbf{E3}  & 8.64  & 53.8 & 28.1 & 790.79 \\
\textbf{E4}  & 11.65 & 57.2 & 29.1 & 848.40 \\
\textbf{E5}  & 10.62 & 60.9 & 31.2 & 971.44 \\ \midrule
\textbf{E6}  & 32.38 & 53.5 & 26.3 & 689.63 \\
\textbf{E7}  & 14.79 & 53.1 & 24.7 & 612.17 \\
\textbf{E8}  & 12.62 & 48.0 & 21.2 & 447.57 \\
\textbf{E9}  & 10.44 & 53.0 & 23.4 & 548.17 \\
\textbf{E10} & 20.06 & 62.3 & 31.3 & 982.65 \\ \bottomrule
\end{tabular}
\end{table}

\begin{figure*}[!t]
    \centering\hfill
    \begin{tabular}{cc}
        \begin{subfigure}[b]{0.45\linewidth}
            \centering
            \includegraphics[width=\linewidth]{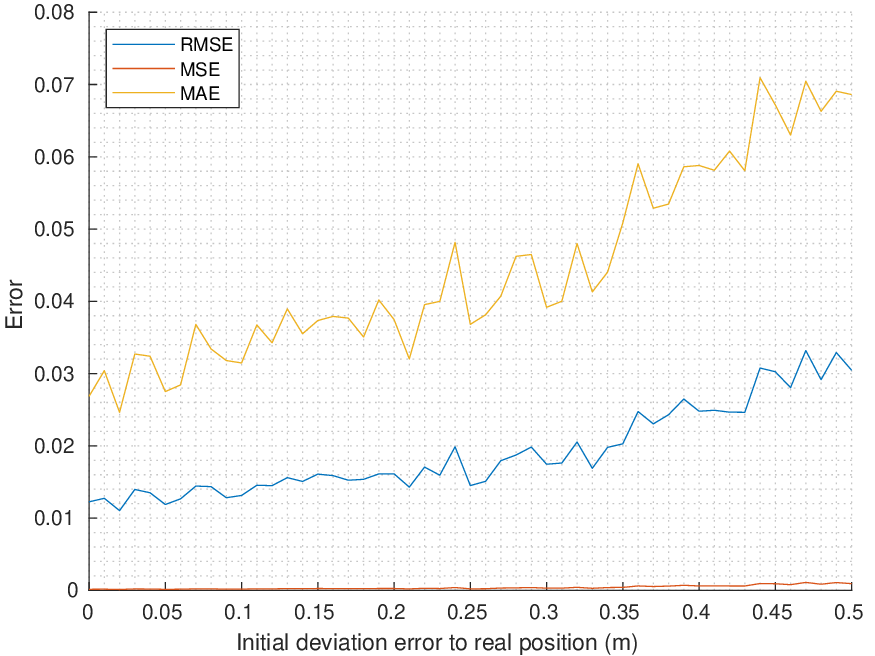}
            \caption{E1}
        \end{subfigure} &
        \begin{subfigure}[b]{0.45\linewidth}
            \centering
            \includegraphics[width=\linewidth]{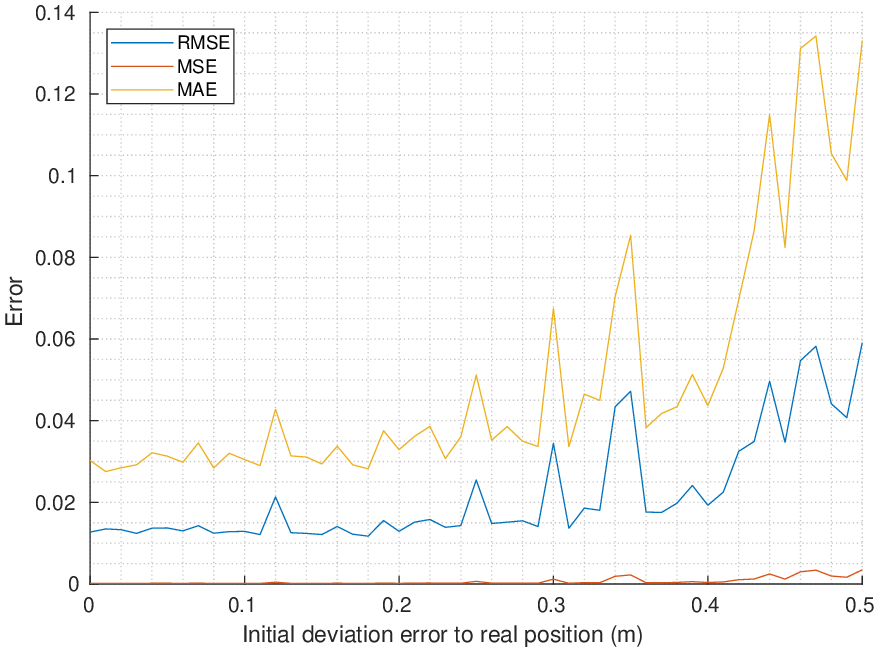}
            \caption{E6}
        \end{subfigure}\hspace{30pt}\\
        \begin{subfigure}[b]{0.45\linewidth}
            \centering
            \includegraphics[width=\linewidth]{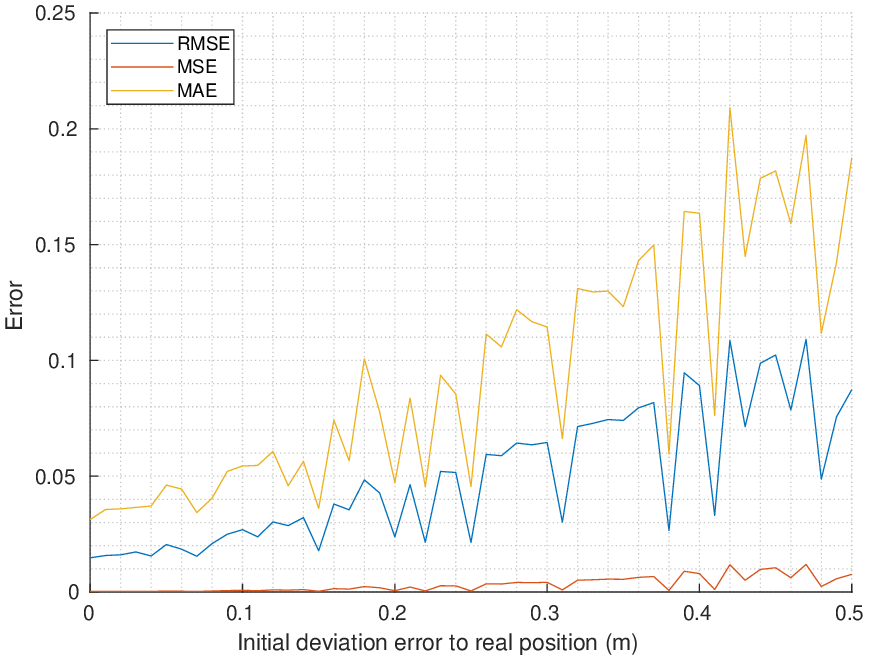}
            \caption{E5}
        \end{subfigure} &
        \begin{subfigure}[b]{0.45\linewidth}
            \centering
            \includegraphics[width=\linewidth]{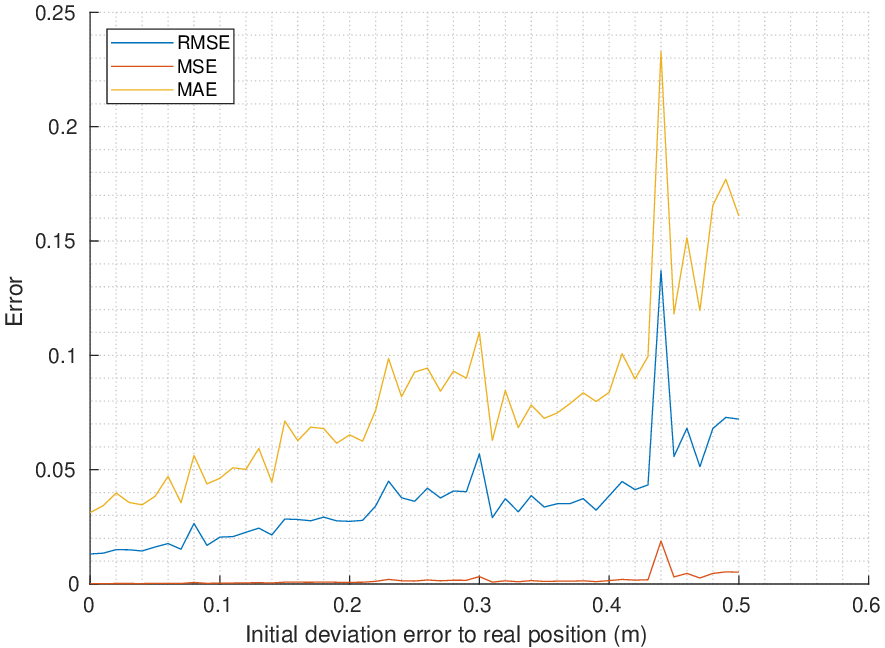}
            \caption{E10}
        \end{subfigure}
    \end{tabular}\hfill\hfill
    \caption{Average error for the recoverability of the loss functions for the \ac{bve} + \ac{ekf} considering different initial estimation errors. Blue -- \ac{rmse}; red -- \ac{mse}; yellow -- \ac{mae}}
    \label{fig: recover-error}
\end{figure*}

Analyzing the table \ref{tab: error bve + ekf}, we verify that simpler and more flexible restrictions result in smaller estimation errors. However, discarding E1, all the experiments conducted in similar estimation errors with an Euclidean error of about \qty{30}{\milli\meter}, if considering the loss function \eqref{eq: loss function}. Despite constraining, the \ac{bve} + \ac{ekf} can perform similarly while increasing the constraining difficulty. Differently, the loss function \ref{eq: loss function v2} has a more progressive behavior, having better results than \eqref{eq: loss function} for less constraining restrictions.

% The algorithm also tends to create circular paths while inferring the objects' position due to its tendency to search for orthogonal observations of the fruit. That performs an interesting behavior that could facilitate the approximation to the fruit while looking for it, with a refined penalization of the loss function that penalizes the distance to the object. 

In a general evaluation, we can conclude that using a differentiable loss function (experiments E1 to E5) such as the dispersion \eqref{eq: loss function} brings advantages over a none differentiable loss function (experiments E6 to E10) such as \eqref{eq: loss function v2}, which depends on a maximum operation. Besides, empirical analysis of the performance of both loss functions under the same conditions concludes that the dispersion loss function was also faster to compute because it has one less inversion matrix to calculate. Simpler and less restrictive conditions deliver lower errors while estimating the position of the fruits once the camera has more freedom to navigate around the region of interest. In both strategies, the \ac{bve} tends to plan an approximated circular path in the case where the algorithm is free to design its path to the most restricted cases (Figures \ref{fig: bve+ekf motion plot}).  These circular paths do not always happen in the same plan but in various plans, even transversal ones.

For some applications, such as precise and careful fruit picking with cutting tools, the reported estimation errors may not be small enough. To improve the error, the system should apply strategies and algorithms that reduce the covariance. Freer systems also prove to have smaller errors because of the liberty of the algorithm to choose the best observation positions. Besides, implementing historical knowledge should optimize innovation and promote the acquisition of new scene perspectives. 

The previous conclusions are enough to understand the performance of both models but do not allow us to understand their limitations and recovery capacity. So, we also performed a recoverability analysis for the loss function to approximate the fruit's position correctly. To achieve this aim, we made multiple essays for estimating the real position of the fruits, considering an initial estimation error between \qtyrange{0}{0.5}{\m} in steps of \SI{1}{\centi\metre}. We considered ten simulations for each initial estimation error and computed the average result. Figure \ref{fig: recover-error} illustrates the average errors, given the initial conditions for experiments E1 and E5. 

From these experiments, we can conclude that both loss functions perform similarly under the most complex and demanding restrictions. Still, the algorithm can tolerate bigger initial estimation errors by using the dispersion minimization loss function  \eqref{eq: loss function}. Besides, this loss function is also easier to compute, and the next viewpoint is estimated quickly and easily. 

% \begin{figure}
%     \centering
%     \begin{subfigure}[b]{0.49\textwidth}
%         \centering
%         \includegraphics[width=\textwidth]{figures/MAPE_E2.1.eps}
%         \caption{E2.1}
%     \end{subfigure}\hfill
%     \begin{subfigure}[b]{0.49\textwidth}
%         \centering
%         \includegraphics[width=\textwidth]{figures/MAPE_E2.6.eps}
%         \caption{E2.6}
%     \end{subfigure}\hfill
%     \begin{subfigure}[b]{0.49\textwidth}
%         \centering
%         \includegraphics[width=\textwidth]{figures/MAPE_E2.5.eps}
%         \caption{E2.5}
%     \end{subfigure}\hfill
%     \begin{subfigure}[b]{0.49\textwidth}
%         \centering
%         \includegraphics[width=\textwidth]{figures/MAPE_E2.10.eps}
%         \caption{E2.10}
%     \end{subfigure}\hfill
%     \caption{Average \ac{mape} for the recoverability of the loss functions for the \ac{bve} + \ac{ekf} considering different initial estimation errors.}
%     \label{fig: recover-mape}
% \end{figure}

\begin{figure*}[!htb]
    \centering
    \hfill
    \begin{subfigure}[b]{0.45\linewidth}
        \centering
        \includegraphics[width=\linewidth]{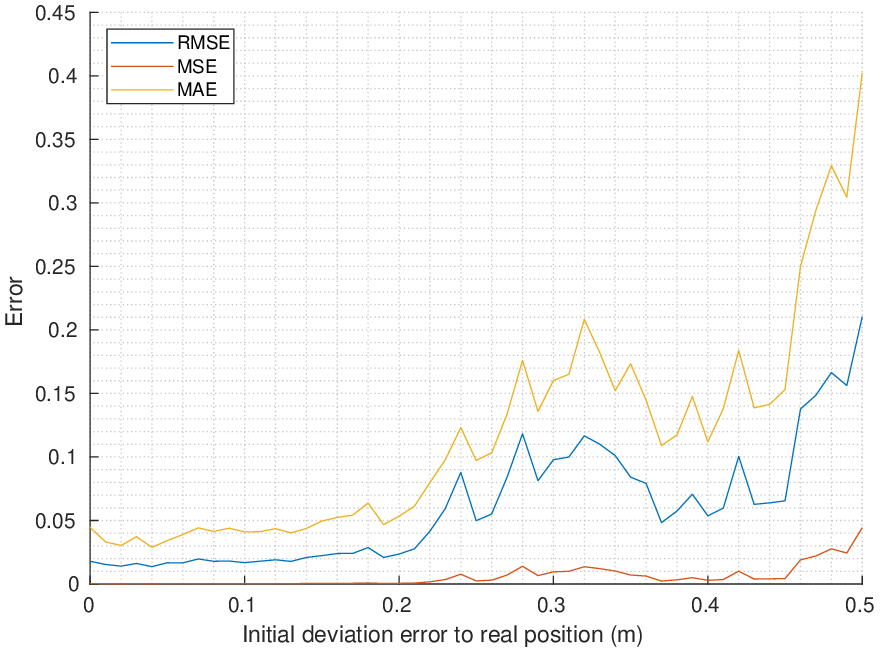}
        \caption{Error}
    \end{subfigure}\hfill
    \begin{subfigure}[b]{0.45\linewidth}
        \centering
        \includegraphics[width=\textwidth]{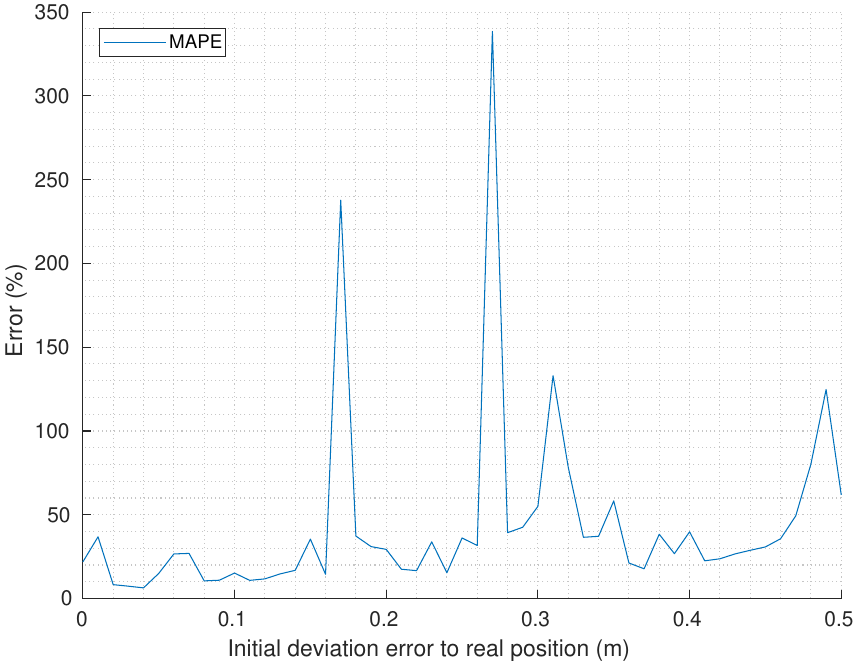}
        \caption{\ac{mape}}
    \end{subfigure}\hfill\hfill
    \caption{Average Error and \ac{mape} for the recoverability of the loss functions for the \ac{bve} + \ac{ekf} considering different initial estimation errors for the loss function \eqref{eq: improved loss function} and the restrictions such as in E5. On left (blue -- \ac{rmse}; red -- \ac{mse}; yellow -- \ac{mae}); On Right (blue -- \ac{mape}).}
    \label{fig: approximation-recover}
\end{figure*}

\begin{figure}[!htb]
    \centering
    \includegraphics[width=0.5\textwidth]{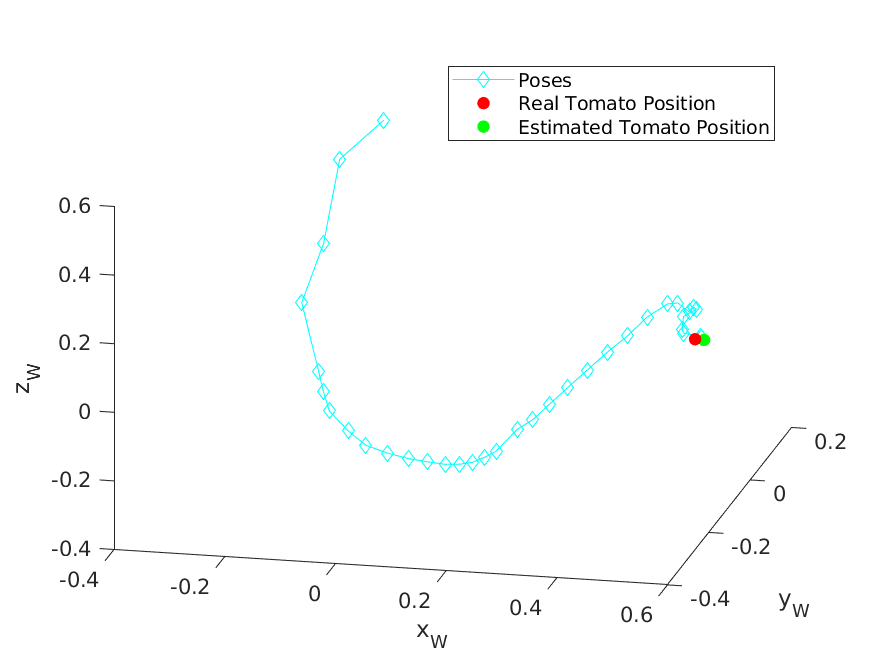}
    \caption{Sensor approximation's path using the loss function \eqref{eq: improved loss function} and the restrictions such as in E5. Light blue -- poses; red -- real fruit position; green -- estimated fruit position.}
    \label{fig: E2.5 approximation}
\end{figure}

As has been observed, the algorithm is effective in searching for the best viewpoint to estimate the position of the fruits. However, task execution is also relevant to ensure the sensory apparatus moves toward the object. The algorithms can approximate the object in a two-step procedure: positioning the fruit and moving to it. However, using a properly designed loss function such as \eqref{eq: improved loss function}, the \ac{bve} + \ac{ekf} algorithms can iteratively refine the fruit's position while moving towards it. Figure \ref{fig: E2.5 approximation} illustrates a possible path to move the sensors from the starting pose to the object, considering the restrictions E5. This scheme shows that the algorithm tends to have a circular path while correcting the fruit's position.

Similar to the previous examples, we also performed a recoverability analysis of the algorithm running the loss function \eqref{eq: improved loss function}. The behavioral results are illustrated in figure \ref{fig: approximation-recover} plots. The algorithm tends to perform worse and accumulate more errors for this function. Here, we can rely on an initial estimation error until about \qty{15}{\centi\metre}. Initial estimation errors bigger than that will result in a final significant estimation error.

% \section{Discussion}
% \label{sec:discussion}

\section{Conclusions}
\label{sec:conclusion}

The robotization of agricultural fields is an approach that can help to overcome some current societal challenges, such as labor shortages in the field or improved crops. However, that requires the implementation of robust \ac{3d} or \ac{6d} perception systems independent of depth sensors because of their sensibility to external perturbances.

To approach the problem, we studied a Gaussian-based solution to minimize the observation covariance over the fruits, which we called \ac{bve}. We powered the \ac{bve} with an \ac{ekf} that iteratively approximates the position of the fruits. The essay was deployed and tested in mathematical simulation over MATLAB. We designed two loss functions to optimize the resulting observability error: a covariance dispersion-based function and the maximum variance of the covariance matrix. The system reached reasonable results with average Euclidean errors lower than \qty{31.2}{\milli\meter}. A more distinctive analysis concludes that the maximum covariance function is more sensitive to restrictions, so a lower error with fewer constraining restrictions. On the other hand, the dispersion-based function is empirically faster to compute and more robust.

Additional evaluations were conducted to assess the algorithm's robustness to different initial conditions, which show that both loss functions perform similarly. A variant loss function that drives the sensor to the object proves the robot can perform both tasks simultaneously. 

Future work should focus on developing the system in a robotic system under controlled environments. Besides using \ac{ekf}, other equivalent algorithms may be tested, such as the Unscented Kalman Filter (UKF). Additional improvements may also be studied, such as implementing historical knowledge that promotes the selection of newer innovative poses.

\section*{Acknowledgements}
\ifdefined\DOUBLEBLIND\else

%This work is co-financed by Component 5 -- Capitalization and Business Innovation, integrated in the Resilience Dimension of the Recovery and Resilience Plan within the scope of the Recovery and Resilience Mechanism (MRR) of the European Union (EU), framed in the Next Generation EU, for the period 2021--2026, within project PhenoBot-LA8.3, with reference PRR-C05-i03-I-000134-LA8.3.

This work is financed by National Funds through the FCT -- Fundação para a Ciência e a Tecnologia, I.P. (Portuguese Foundation for Science and Technology) within the project OmicBots, with reference PTDC/ASP-HOR/1338/2021 (\href{http://dx.doi.org/10.54499/PTDC/ASP-HOR/1338/2021}{DOI:10.54499/PTDC/ASP-HOR/1338/2021}).

% This work is co-financed by Component 5 -- Capitalization and Business Innovation, integrated in the Resilience Dimension of the RRP within the scope of the MRR of the EU, framed in the Next Generation EU, for the period 2021--2026, within project PhenoBot-LA8.3, with reference PRR-C05-i03-I-000134-LA8.3.

Sandro Costa Magalhães is granted by the Portuguese Foundation for Science and Technology (FCT) through the ESF integrated into NORTE2020, under scholarship agreement SFRH/BD/147117/2019 (\href{http://dx.doi.org/10.54499/SFRH/BD/147117/2019}{DOI:10.54499/SFRH/BD/147117/2019}).

\fi

\bibliographystyle{apalike}
{\small
\bibliography{Example}}
%\printbibliography[category=inbib]}

%\section*{{Appendix}}

\end{document}